\documentclass[11pt]{article}
\usepackage{microtype}
\usepackage[margin=1in]{geometry}
\usepackage{amsmath, amssymb}
\usepackage{graphicx}
\usepackage[numbers]{natbib}
\usepackage{booktabs}
\usepackage{hyperref}
\usepackage{float}      
\usepackage{placeins}   
\usepackage{caption}
\usepackage{subcaption}
\usepackage{microtype}
\graphicspath{{samples/}}  
\usepackage{needspace}    

\title{CADE 2.5: ZeResFDG --- Frequency-Decoupled, Rescaled and Zero-Projected Guidance for SD/SDXL Latent Diffusion Models}
\author{
Denis Rychkovskiy ("DZRobo", Independent Researcher)\\
\texttt {nebularus@yandex.ru}
}
\date{October 11, 2025}

\begin{document}
\maketitle
\vspace{-0.5em}
\begin{center}
{\large Primary Subject Class: cs.CV \quad Secondary Class: cs.LG}
\end{center}

\begin{abstract}
We introduce \textbf{CADE 2.5} (Comfy Adaptive Detail Enhancer), a sampler-level guidance stack for \textbf{SD/SDXL latent diffusion models}. 
The central module, \textbf{ZeResFDG}, unifies (i) frequency-decoupled guidance that reweights low- and high-frequency components of the guidance signal, 
(ii) energy rescaling that matches the per-sample magnitude of the guided prediction to the positive branch, and 
(iii) zero-projection that removes the component parallel to the unconditional direction. 
A lightweight spectral EMA with hysteresis switches between a conservative and a detail-seeking mode as structure crystallizes during sampling. 
Across SD/SDXL samplers, ZeResFDG improves sharpness, prompt adherence, and artifact control at moderate guidance scales \emph{without any retraining}. 
In addition, we employ a training-free inference-time stabilizer, \textbf{QSilk Micrograin Stabilizer} (quantile clamp + depth/edge-gated micro-detail injection), which improves robustness and yields natural high-frequency micro-texture at high resolutions with negligible overhead.
For completeness we note that the same rule is compatible with alternative parameterizations (e.g., velocity), which we briefly discuss in the Appendix; however, this paper focuses on SD/SDXL latent diffusion models.
\end{abstract}

\section{Introduction}
Latent diffusion models (e.g., SD/SDXL) deliver high-fidelity image synthesis but can degrade at large classifier-free guidance (CFG) scales, 
exhibiting oversaturation, tone drift, or texture artifacts~\citep{sadat2024oversaturation}. 
Reducing CFG to avoid these effects often sacrifices sharpness and prompt adherence. 
Prior work addresses the trade-off via attention-based guidance (e.g., SAG/PAG)~\citep{hong2022sag, ahn2024pag}, 
schedule-aware or interval-limited guidance~\citep{kynkaanniemi2024limited}, and rescaling heuristics widely used in practice~\citep{diffusersPix2Pix}.

We propose a compact sampler-side stack called \textbf{CADE 2.5}. 
Its core, \textbf{ZeResFDG}, re-shapes the guidance itself by combining: 
(1) \emph{frequency decoupling} to protect global tone/structure while selectively enhancing micro-detail; 
(2) \emph{energy rescaling} to mitigate overexposure at high CFG; and 
(3) \emph{zero-projection} to suppress early-step drift along the unconditional direction.
A tiny spectral EMA with hysteresis toggles between a conservative and a detail-seeking mode during sampling.

Our work is complementary to the Adaptive Projected Guidance (APG) framework by Sadat et al. (2025)~\citep{sadat2025apg}, which decomposes classifier-free guidance into parallel and orthogonal components; we extend this perspective with rescaling and a zero-projection term specifically tailored for SD/SDXL latent diffusion.

\section{Background}
\paragraph{Classifier-free guidance (CFG).}
Given conditional and unconditional predictions $(y_c, y_u)$ at the same latent state, standard CFG forms 
$y_{\text{cfg}} = y_u + s\,(y_c - y_u)$ with scale $s > 0$. 
Large $s$ often yields color blowouts and haloing~\citep{sadat2024oversaturation}. 
Attention-oriented control (SAG/PAG)~\citep{hong2022sag, ahn2024pag} and \emph{limited-interval} application of guidance~\citep{kynkaanniemi2024limited} suppress artifacts, 
while practical pipelines frequently apply a \emph{guidance rescale} to match energies of branches~\citep{diffusersPix2Pix}.

\section{Method}
Let the model output $y$ in the standard $\varepsilon$-parameterization used by SD/SDXL samplers. 
For $(y_c, y_u)$, define the raw guidance $\Delta = y_c - y_u$.

\paragraph{Frequency-Decoupled Guidance (FDG)~\citep{sadat2025fdg}.}
We split $\Delta$ into low/high bands via a Gaussian low-pass $G_\sigma$: 
$\Delta_\ell = G_\sigma * \Delta$, $\Delta_h = \Delta - \Delta_\ell$, 
and reweight them as 
$\tilde{\Delta} = \lambda_\ell \Delta_\ell + \lambda_h \Delta_h$, with $\lambda_\ell \in [0,1]$, $\lambda_h \gtrsim 1$.

\paragraph{RescaleCFG (energy match).}
We form $y_{\text{cfg}} = y_u + s\,\tilde{\Delta}$ and rescale it to match the per-sample standard deviation of $y_c$, then blend with a coefficient $\alpha \in [0,1]$:
\begin{equation}
y_{\text{res}} = \alpha \cdot \mathrm{Rescale}\!\big(y_{\text{cfg}}, \mathrm{std}(y_c)\big) + (1-\alpha)\,y_{\text{cfg}}.
\end{equation}

\paragraph{Zero-Projection (CFGZero).}
To suppress leakage along the unconditional direction, compute 
$\alpha_\parallel = \langle y_c, y_u\rangle / \langle y_u, y_u\rangle$ 
and use the residual $r = y_c - \alpha_\parallel y_u$ as the signal to guide (optionally FDG-filtered).
\textbf{Relation to prior work.} Our formulation conceptually aligns with the
projection analysis of classifier-free guidance proposed by Sadat et al.\,(2025),
who demonstrated that down-weighting the parallel component mitigates
oversaturation effects in diffusion models~\citep{sadat2025apg}.

\paragraph{Spectral controller (EMA + hysteresis).}
We monitor a high-frequency ratio $r_{\mathrm{HF}} = \|\Delta_h\|^2/(\|\Delta_\ell\|^2+\|\Delta_h\|^2)$ and track an EMA $\rho$. 
With two thresholds $(\tau_{\mathrm{lo}}, \tau_{\mathrm{hi}})$ and hysteresis, we switch between the conservative mode (\emph{CFGZeroFD}) and the detail-seeking mode (\emph{RescaleFDG}).

\paragraph{Auxiliary stabilizers.}
We employ a small attention normalization patch (NAG\cite{chen2025normalizedattentionguidanceuniversal}) in the positive branch, optional local spatial gating from external masks (e.g., faces/hands), 
a tiny early-step exposure-bias scale, and a directional post-mix (Muse Blend). 
All components are training-free and implemented as a sampler wrapper for SD/SDXL pipelines.

\subsection{Inference-Time Stabilizers: QSilk Micrograin Stabilizer}
We complement ZeResFDG with a lightweight, training-free stabilizer that acts during inference and requires no changes to model weights.

\paragraph{Per-step quantile clamp (QClamp).}
After each denoising step $i$, we apply a per-sample quantile clamp to the denoised tensor, clipping values to the $(0.1\% , 99.9\%)$ percentiles computed per sample. This softly removes rare value spikes and prevents NaN/Inf cascades with negligible overhead.

\paragraph{Late-tail micro-detail injection (depth/edge-gated).}
On late steps (tail of the sigma schedule), we add a tiny high-frequency residual in image space, gated by both edges and depth to avoid halos and to favor near surfaces:
\begin{equation}
\label{eq:microinject}
x'_{\mathrm{img}} \;=\; x_{\mathrm{img}} \;+
\alpha(t)\, g_{\mathrm{edge}}\, g_{\mathrm{depth}}\, \big( x_{\mathrm{img}} - G_{\sigma}(x_{\mathrm{img}}) \big),
\end{equation}
where $G_{\sigma}$ is a small Gaussian blur (fine-scale high-pass), $g_{\mathrm{edge}}$ is an inverse Sobel-magnitude gate to suppress sharpening near strong edges, and $g_{\mathrm{depth}}$ is a normalized depth gate (favoring nearer surfaces). The scalar $\alpha(t)$ smoothly ramps up only near the end of the schedule. In practice this produces realistic micro-texture (pores, peach fuzz) at $4\mathrm{K}$--$6\mathrm{K}$ without oversharpening.

Both components are implementation choices that remain \emph{orthogonal} to ZeResFDG and other guidance rules; they are training-free and add only a small constant overhead at inference time.

\section{Algorithm}
\begin{figure}[H]
\hrule
\vspace{0.5em}
\textbf{Algorithm 1: ZeResFDG (per step; SD/SDXL, $\varepsilon$-parameterization)}\\[0.25em]
\hrule
\vspace{0.5em}
\small
\begin{enumerate}
\item Inputs: $y_c, y_u$ (cond/uncond), guidance $s$, rescale mix $\alpha$, FDG gains $(\lambda_\ell,\lambda_h)$, thresholds $(\tau_{\mathrm{lo}},\tau_{\mathrm{hi}})$, EMA $\rho$, optional spatial mask $g(x,y)$.
\item $\Delta \leftarrow y_c - y_u$; \quad $\Delta_\ell \leftarrow G_\sigma * \Delta$; \quad $\Delta_h \leftarrow \Delta - \Delta_\ell$.
\item Update $r_{\mathrm{HF}} = \|\Delta_h\|^2/(\|\Delta_\ell\|^2+\|\Delta_h\|^2)$ and EMA $\rho$; set mode $\in \{\text{CFGZeroFD}, \text{RescaleFDG}\}$ via hysteresis on $\rho$.
\item \textbf{If} mode = CFGZeroFD: 
  \begin{enumerate}
  \item $\alpha_\parallel \leftarrow \langle y_c, y_u\rangle / \langle y_u, y_u\rangle$; \quad $r \leftarrow y_c - \alpha_\parallel y_u$.
  \item $\tilde{\Delta} \leftarrow \lambda_\ell (G_\sigma * r) + \lambda_h (r - G_\sigma * r)$.
  \item \textbf{If} mask $g$: $\tilde{\Delta} \leftarrow g \cdot \tilde{\Delta}$.
  \item $y \leftarrow \alpha_\parallel y_u + s \cdot \tilde{\Delta}$.
  \end{enumerate}
\item \textbf{Else} (RescaleFDG):
  \begin{enumerate}
  \item $\tilde{\Delta} \leftarrow \lambda_\ell \Delta_\ell + \lambda_h \Delta_h$; \quad \textbf{If} mask $g$: $\tilde{\Delta} \leftarrow g \cdot \tilde{\Delta}$.
  \item $y_{\text{cfg}} \leftarrow y_u + s \cdot \tilde{\Delta}$.
  \item $y \leftarrow \alpha \cdot \mathrm{Rescale}\!\big(y_{\text{cfg}}, \mathrm{std}(y_c)\big) + (1-\alpha)\,y_{\text{cfg}}$.
  \end{enumerate}
\item Return $y$.
\end{enumerate}
\normalsize
\vspace{-0.25em}\hrule
\end{figure}

\section{Implementation details}
\textbf{Defaults.} We use $\sigma{=}1.0$ for the Gaussian split, $(\lambda_\ell,\lambda_h){=}(0.6,1.3)$, rescale mix $\alpha{=}0.7$, 
EMA $\beta{=}0.8$, hysteresis thresholds $(\tau_{\mathrm{lo}},\tau_{\mathrm{hi}}){=}(0.45,0.60)$; 
NAG\cite{chen2025normalizedattentionguidanceuniversal} on the positive branch; optional local masks for faces/hands; and a small early-step exposure-bias scale. 
\textbf{Integration.} The stack is a training-free sampler wrapper and fits SD/SDXL pipelines (e.g., ComfyUI nodes).

\FloatBarrier               
\Needspace*{0.82\textheight}
\section{Visual Results}
Qualitative examples illustrating typical gains on portraits (eyes, hair, skin) and challenging hand regions (fingers, nails).

\begin{figure}[H]
  \centering
  \begin{subfigure}[b]{0.48\linewidth}
    \centering
    \includegraphics[width=\linewidth]{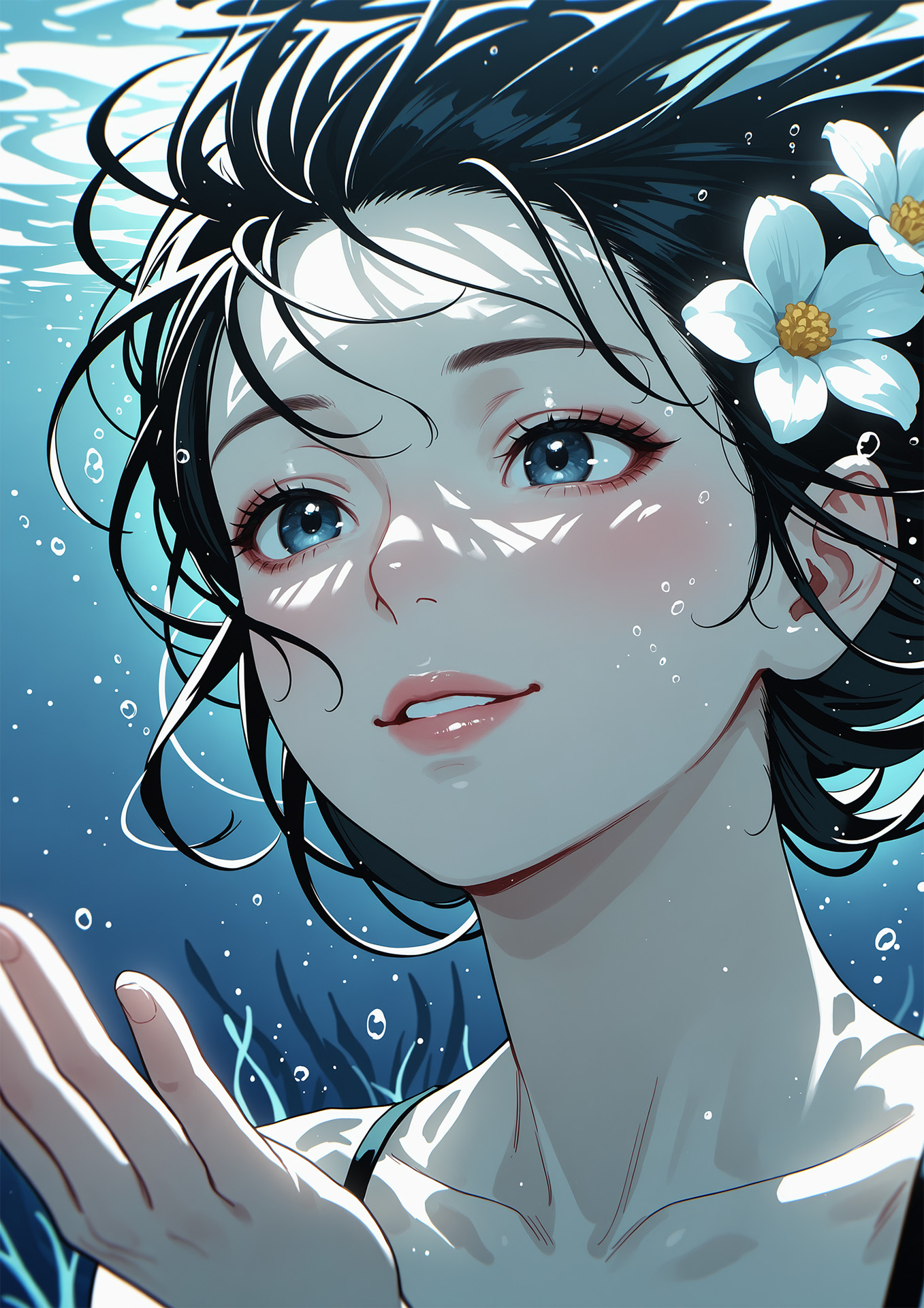}
    \caption*{\textbf{Anime Portrait} --- Cleanest result. Enhancing lines, colors and light.}
  \end{subfigure}\hfill
  \begin{subfigure}[b]{0.48\linewidth}
    \centering
    \includegraphics[width=\linewidth]{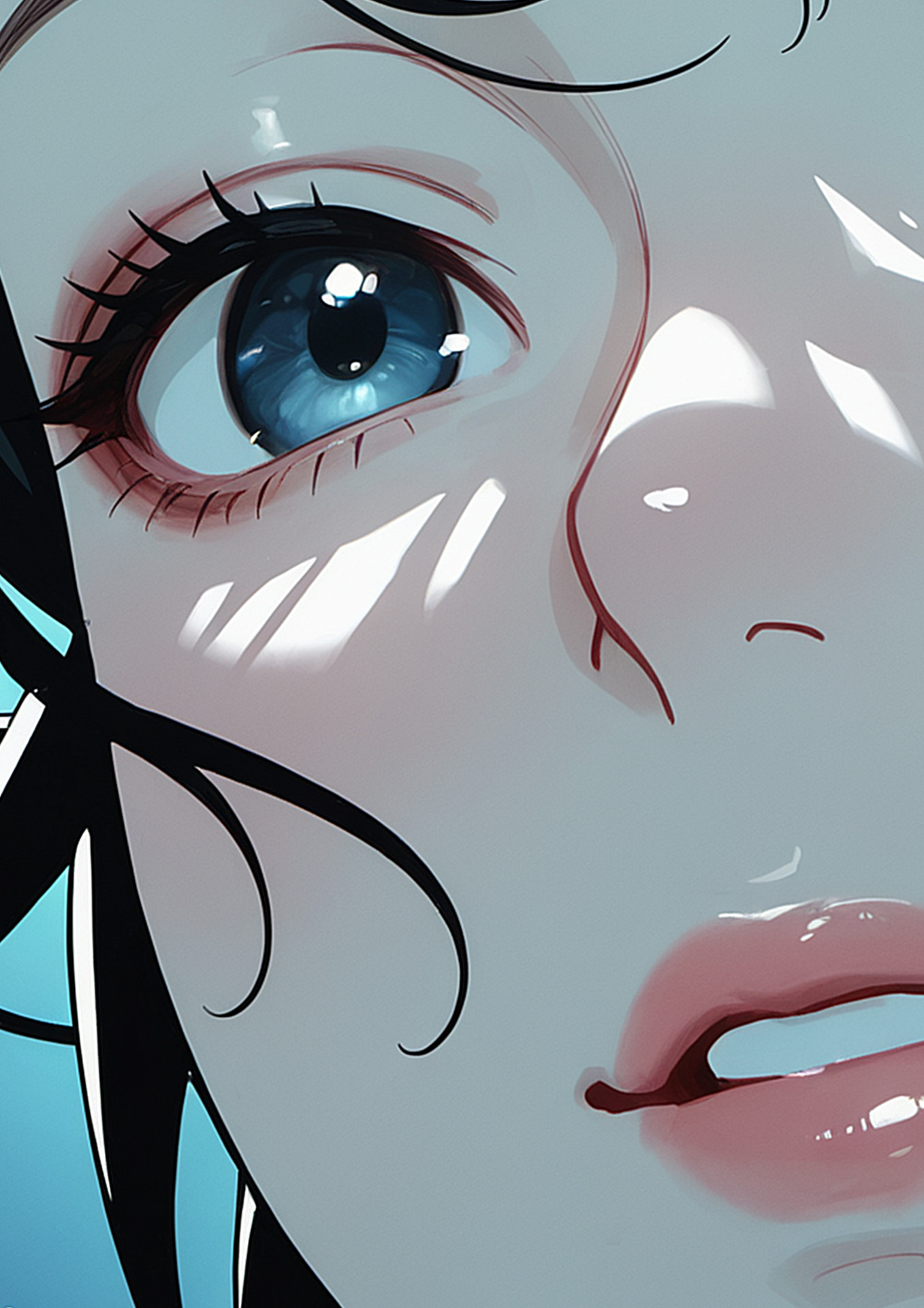}
    \caption*{\textbf{Crop: Eye, Nose, Lips} --- amazing lines and zero jitter.}
  \end{subfigure}
  \vspace{0.25em}
  \caption{Qualitative samples "Anime style" produced with CADE 2.5 (ZeResFDG pipe (SDXL)).}
  \label{fig:1}
\end{figure}

\begin{figure}[H]
  \centering
  \begin{subfigure}[b]{0.48\linewidth}
    \centering
    \includegraphics[width=\linewidth]{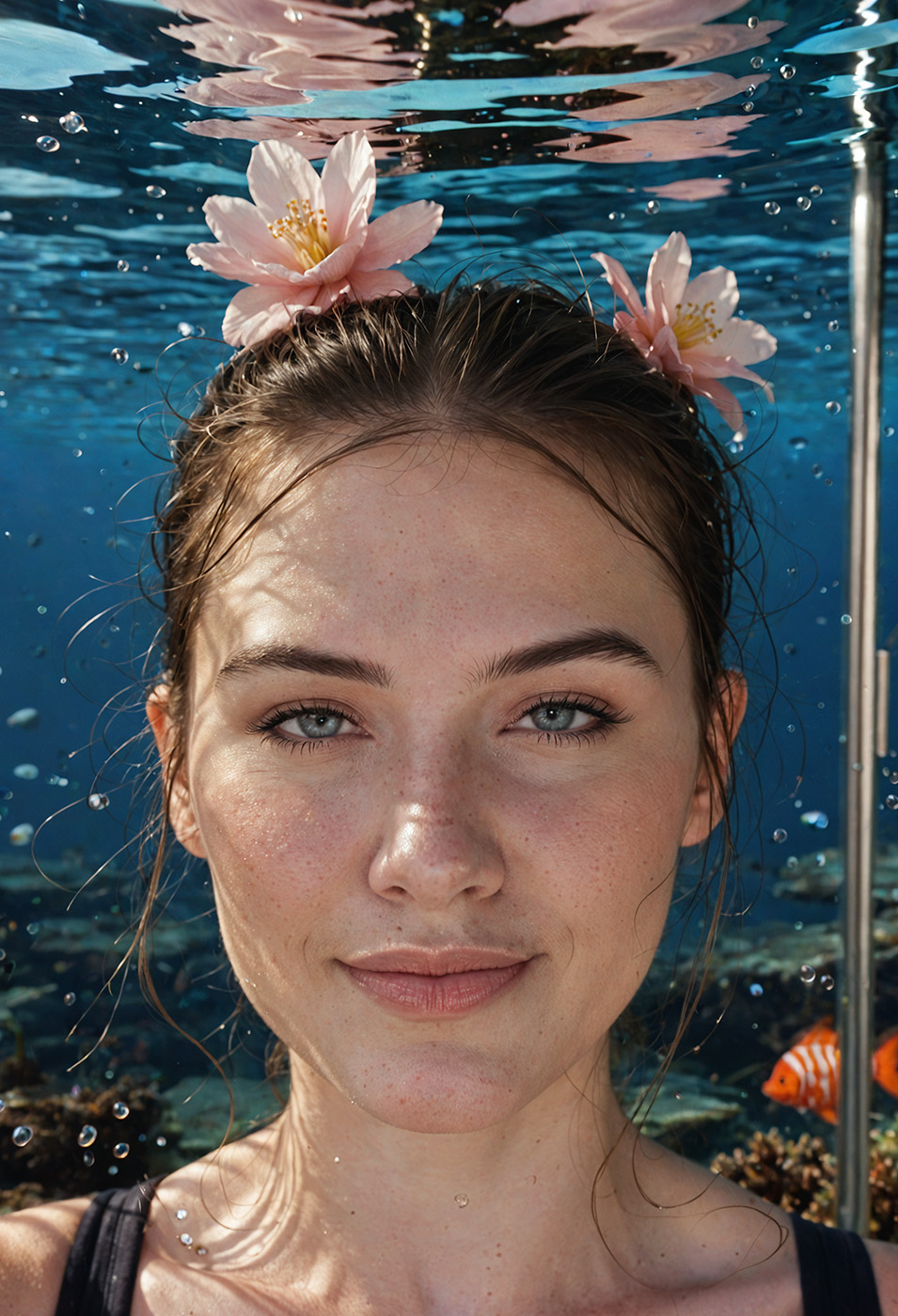}
    \caption*{\textbf{Photo Portrait} --- ZeResFDG preserves global tone while enhancing micro-detail.}
  \end{subfigure}\hfill
  \begin{subfigure}[b]{0.48\linewidth}
    \centering
    \includegraphics[width=\linewidth]{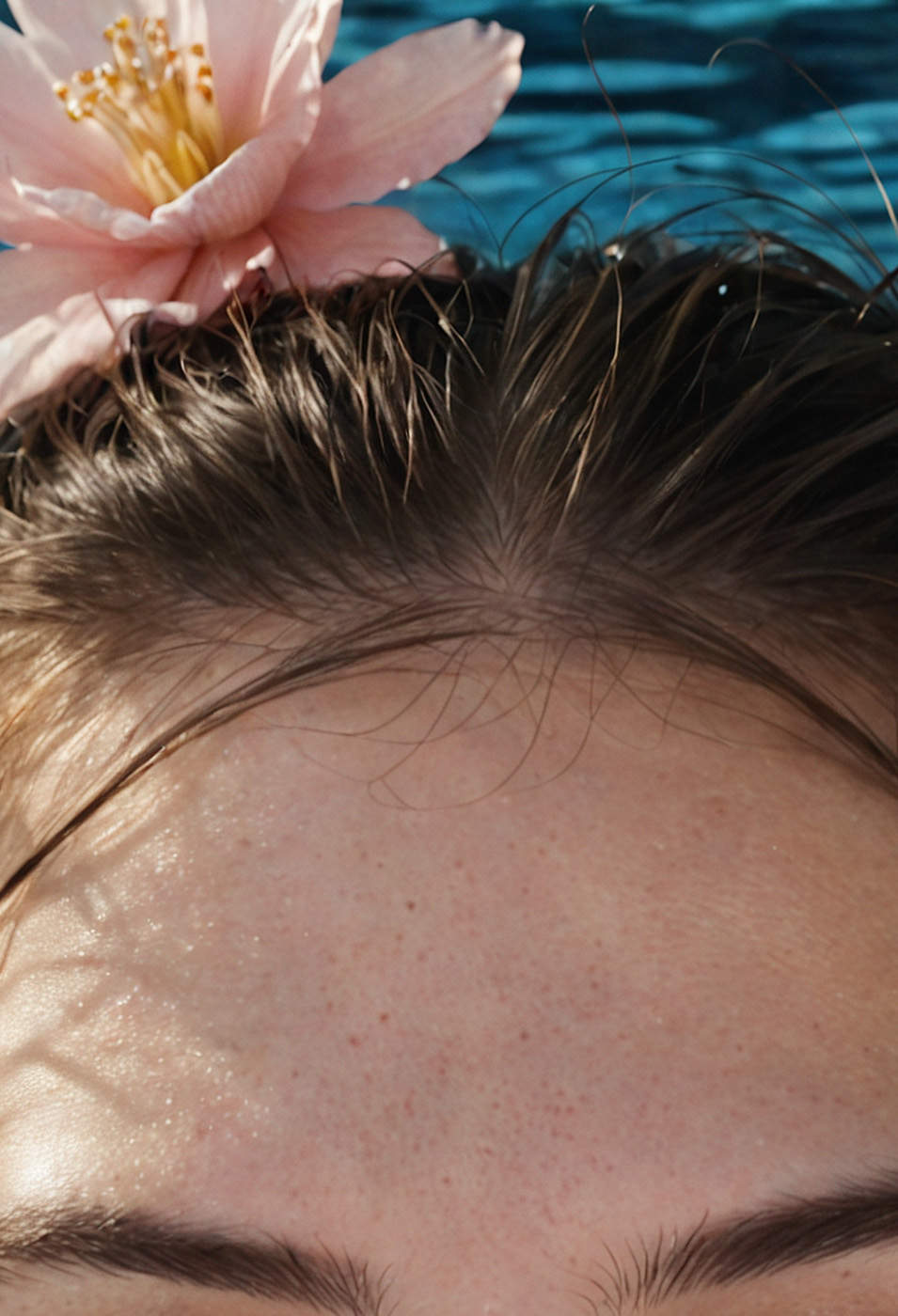}
    \caption*{\textbf{Crop, Face and Hair} --- fewer artifacts in eye, beautiful hair details, skin tone and micro-detail.}
  \end{subfigure}
  \vspace{0.25em}
  \caption{Qualitative samples "Photo style" produced with CADE 2.5 (ZeResFDG pipe (SDXL)).}
  \label{fig:2}
\end{figure}

\begin{figure}[H]
  \centering
  \begin{subfigure}[b]{0.48\linewidth}
    \centering
    \includegraphics[width=\linewidth]{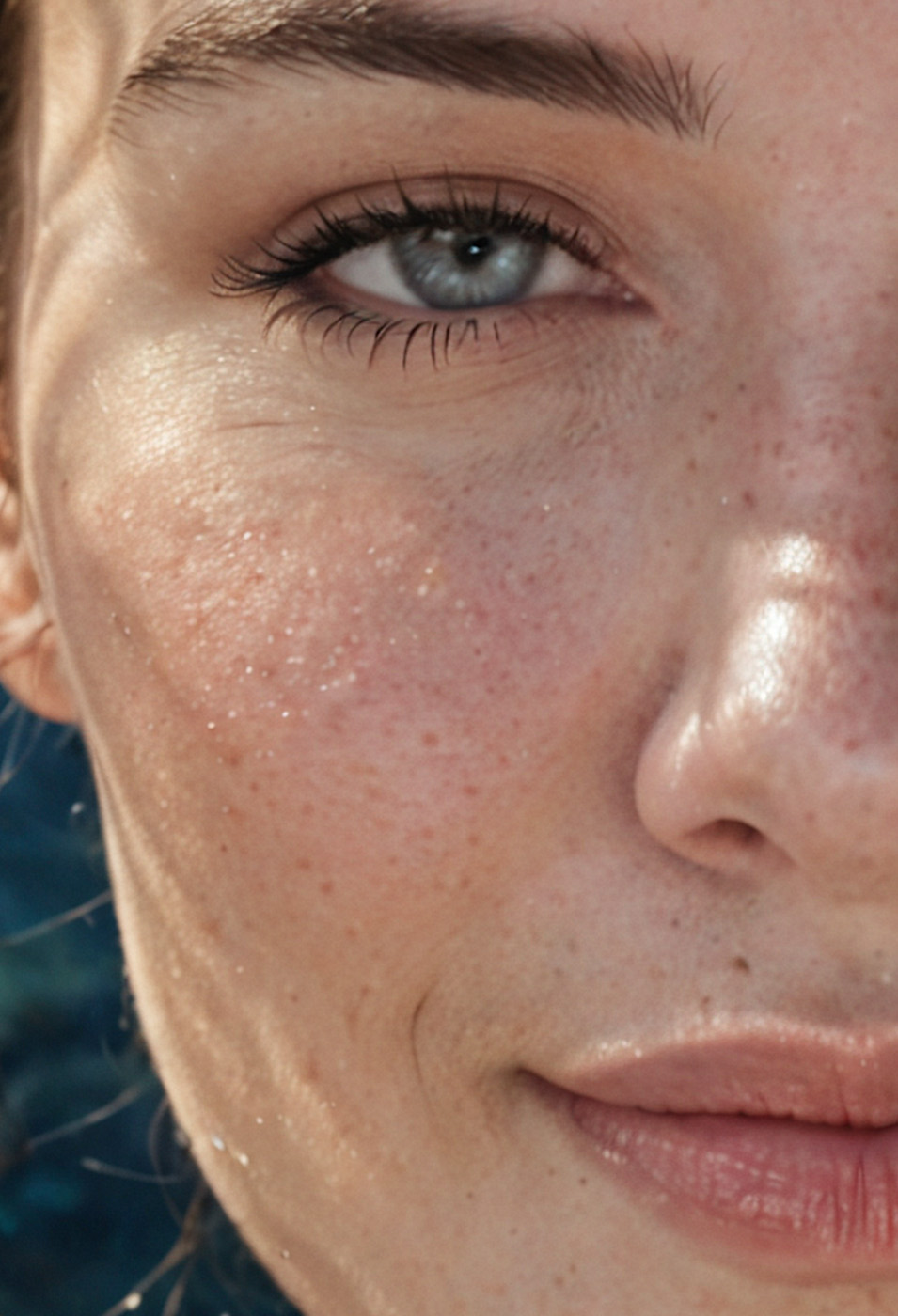}
    \caption*{\textbf{Crop: Lips and Nose} --- enhancing micro-detail.}
  \end{subfigure}\hfill
  \begin{subfigure}[b]{0.48\linewidth}
    \centering
    \includegraphics[width=\linewidth]{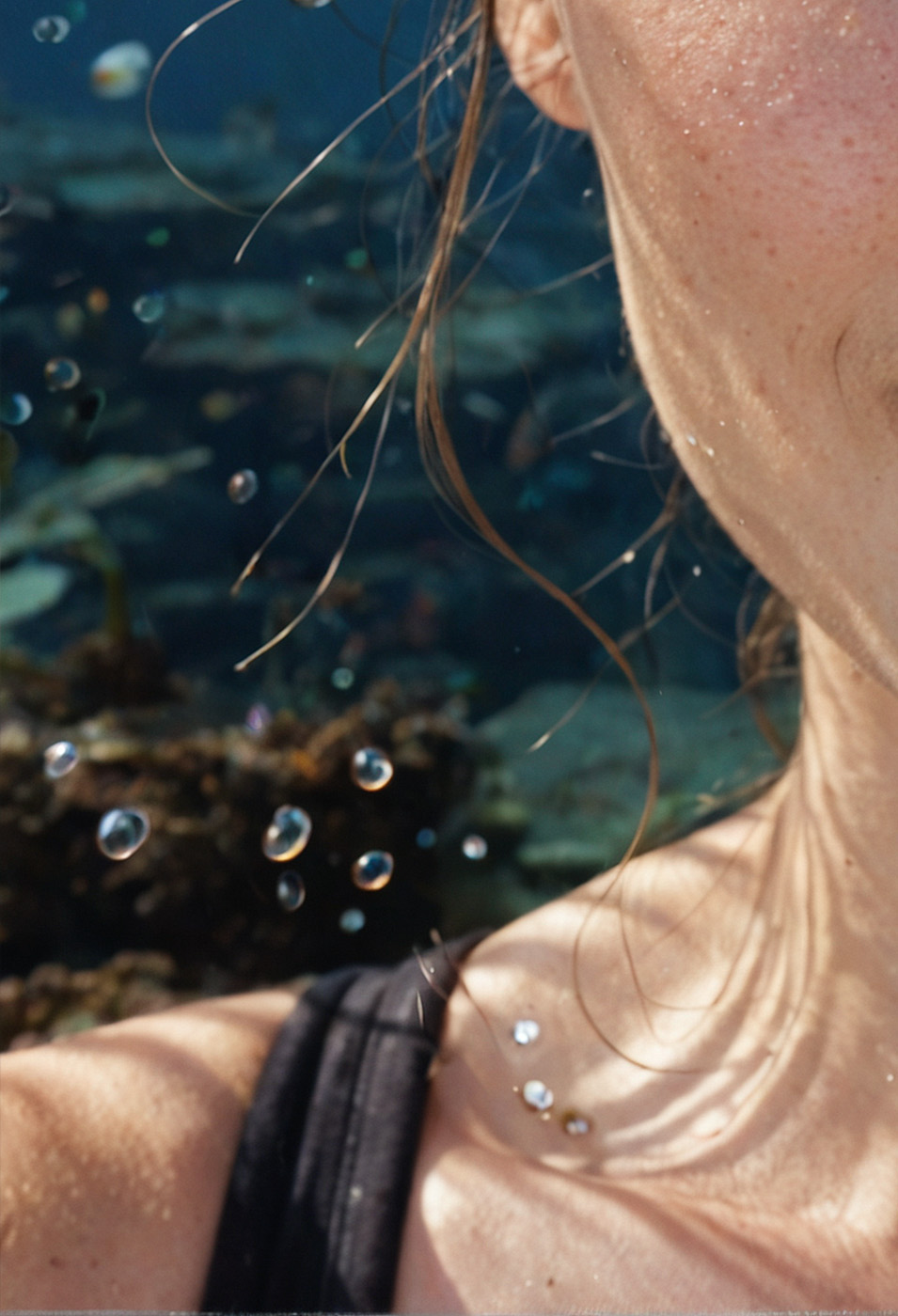}
    \caption*{\textbf{Crop: Neck} --- enhancing micro-detail.}
  \end{subfigure}
  \vspace{0.25em}
  \caption{Qualitative samples "Photo style" produced with CADE 2.5 (ZeResFDG pipe (SDXL)).}
  \label{fig:3}
\end{figure}

\FloatBarrier

\section{Evaluation}
Our goal is to assess \emph{practical sampling behavior} of SD/SDXL pipelines with ZeResFDG under realistic settings.

\paragraph{Setup.} 
We use SDXL with a resolution of $672{\times}944$, a standard sampler (Euler (for Anime)/UniPC (for Photo)) and the same hints for all methods. Each experiment went through 4 consecutive steps through CADE 2.5 (with ZeResFDG enabled), after which the final output image resolution was $3688{\times}5192$ . Our settings: steps - 25, cfg - 4.5, denoise - 0.65. We include the same VAE/text encoders and only change SDXL models (Photo/Anime oriented). 

\paragraph{Generation quality.}
We present images (i) portraits (eyes, hair,
skin tones), (ii) hand (fingers/nails), and (iii) high-frequency textures (human skin). Across these cases, CADE 2.5 (ZeResFDG) maintains global tone and composition while improving
micro-detail and reducing typical high-CFG artifacts (oversaturation, haloing).
Representative examples are shown in Fig. ~\ref{fig:1}, Fig. ~\ref{fig:2}, Fig. ~\ref{fig:3}; extended grids with fixed seeds are included in the supplementary.

\section{Limitations}
Our evaluation is intentionally compact and largely qualitative. 
We focus on typical user settings rather than exhaustive benchmarks; comprehensive distributional metrics and ablations across datasets are left for future work.

\section{Conclusion}
\paragraph{Beyond ZeResFDG (engineering note).}
While this paper focuses on ZeResFDG as the central guidance rule for SD/SDXL, the released CADE node ships with an extended
training-free stack that we found helpful across diverse prompts. In practice we use a \textbf{four-pass preset}:

\begin{itemize}
\item \textbf{Pass I” --- Robust start (early steps).} ZeResFDG with a small exposure-bias scale (EPS), plus a lightweight attention
normalization patch (NAG) on the positive branch. Goal: stabilize tone/structure and suppress early drift.
\item \textbf{Pass II” --- Detail growth (mid steps).} Enable optional local spatial gating
(e.g., CLIPSeg/ONNX masks for faces/hands/pose). Goal: sharpen high-frequency detail while protecting sensitive regions.
\item \textbf{Pass III” --- Balance and finish (late steps).} Keep ZeResFDG and apply a directional post-mix (Muse Blend) with energy matching.
Goal: crisp micro-detail without oversharpening or saturation.
\item \textbf{Pass IV” --- Polish (final touch).} A light polish that preserves low-frequency shape while allowing gentle high-frequency clean-up.
\end{itemize}

These components are implementation choices rather than a new learning objective; they keep the method training-free and add only a small constant overhead.
A thorough ablation of each component is left for future work, and the open-source node exposes all toggles and presets for reproducibility.
\footnote{Implementation is available in the CADE~2.5 node; see code release for details.}

\noindent\textit{Inference-time stabilizer (QSilk Micrograin Stabilizer).}
In addition to ZeResFDG, our public node employs a training-free stabilizer that combines per-step quantile clamp with a depth/edge-gated micro-detail injection on the schedule tail (Eq.~\ref{eq:microinject}). We observe improved robustness and more natural micro-texture at high output resolutions with negligible overhead.

\section*{Acknowledgments}
The author used GPT-5 to assist with drafting, editing, code suggestions, and figure layout. All technical decisions, implementations, experiments, and validation were performed by the human author, who takes full responsibility for the content.

\appendix
\section{Compatibility with Alternative Parameterizations}
While this paper focuses on the standard $\varepsilon$-parameterization in SD/SDXL, the ZeResFDG rule operates identically in velocity space by replacing $(y_c,y_u)$ with $(v_c,v_u)$ 
and forming $v_{\text{cfg}} = v_u + s (v_c - v_u)$ before applying the same zero-projection, FDG~\citep{sadat2025fdg}, and rescaling. 
A thorough study of velocity-parameterized students is left for future work.

\FloatBarrier               
\Needspace*{0.82\textheight}
\bibliographystyle{plainnat}
\bibliography{references}

\end{document}